\documentclass[letterpaper]{article}

\usepackage{times}
\usepackage{epsfig}
\usepackage{graphicx}
\usepackage{amsmath}
\usepackage{amssymb}
\usepackage{commath}
\usepackage{multirow}
\usepackage{wrapfig}
\usepackage{enumitem}
\usepackage[twoside,a4paper,inner=20mm,outer=20mm,top=20mm,bottom=30mm,includehead,heightrounded]{geometry}
\usepackage{floatrow}
\usepackage[pagebackref=true,breaklinks=true,colorlinks,bookmarks=false]{hyperref}
\newfloatcommand{capbtabbox}{table}[][\FBwidth]

\usepackage{xcolor}

\usepackage[utf8]{inputenc}

\title{ISyNet: Convolutional Neural Networks design for AI accelerator}

\author {
    Alexey Letunovskiy$^{1}$\\
    {\tt\small letunovskiy.alexey@huawei.com}\\
    \and
    Vladimir Korviakov$^{1}$\\
    {\tt\small korviakov.vladimir1@huawei.com}\\
    \and
    Vladimir Polovnikov$^{2}$\\
    {\tt\small vladimir.polovnikov@math.msu.ru}\\
    \and
    Anastasiia Kargapoltseva$^{2}$\\
    {\tt\small nastyakargapoltseva1996@gmail.com}\\
    \and
    Ivan Mazurenko$^{1,2}$\\
    {\tt\small mazurenko.ivan1@huawei.com}\\
    \and
    Yepan Xiong$^{1}$\\
    {\tt\small xiongyepan@huawei.com}\\
    \and
    $^{1}$ Intelligent systems and Data science Technology center\\
    Huawei Technologies Co., Ltd\\
    Moscow, Russia\\
    $^{2}$ Lomonosov Moscow State University,\\
    Moscow, Russia
}

\begin{document}

\maketitle

\begin{abstract}
	In recent years Deep Learning reached significant results in many practical problems, such as computer vision, natural language processing, speech recognition and many others. For many years the main goal of the research was to improve the quality of models, even if the complexity was impractically high. However, for the production solutions, which often require real-time work, the latency of the model plays a very important role. Current state-of-the-art architectures are found with neural architecture search (NAS) taking model complexity into account. However, designing of the search space suitable for specific hardware is still a challenging task. To address this problem we propose a measure of hardware efficiency of neural architecture search space -- matrix efficiency measure (MEM); a search space comprising of hardware-efficient operations; a latency-aware scaling method; and ISyNet -- a set of architectures designed to be fast on the specialized neural processing unit (NPU) hardware and accurate at the same time. We show the advantage of the designed architectures for the NPU devices on ImageNet (Figure~\ref{fig:isynet_imagenet}) and the generalization ability for the downstream classification and detection tasks.
\end{abstract}

\section{Introduction}
\label{sec:intro}
Specialized neural processing unit (NPU) hardware was developed for fast neural network inference and training. The core feature of these devices is a design of highly optimized matrix multiplication unit, as the most computationally expensive operations -- convolution and fully-connected layer -- can be reduced to matrix multiplication. Speed of deep neural networks (DNN) inference and training on NPU devices depends not only on amount of matrix multiplication operations, but also on optimal data transfer and vector operations. To design a set of architectures fast for specific device and accurate at the same time a researcher needs to pass the following steps: find and construct a search space, adapt search method for the developed search space, decide how to take complexity of architectures into account. To address these issues we propose several steps, which make process of architectures search much easier.
Our main contributions presented in this paper are:

\begin{itemize}
\item matrix efficiency measure (MEM) - novel measure for numerical evaluation of efficiency of DNN architectures and search spaces to the NPU hardware;
\item NPU-efficient search space design having high MEM value;
\item NPU-efficient scaling algorithm that allows to scale architectures with respect to the latency on the target hardware;
\item ISyNet - novel family of NPU-efficient architectures that provide optimal accuracy vs. latency trade-off and have strong generalization ability proven on various downstream datasets and computer vision tasks. We open-sourced ISyNet architectures in MindSpore Model Zoo.\\ \hyperlink{https://gitee.com/mindspore/models/tree/master/research/cv/ISyNet}{https://gitee.com/mindspore/models/tree/master/research/cv/ISyNet} 
\end{itemize}

\begin{wrapfigure}[27]{r}{75mm}
\includegraphics[width=1.0\linewidth]{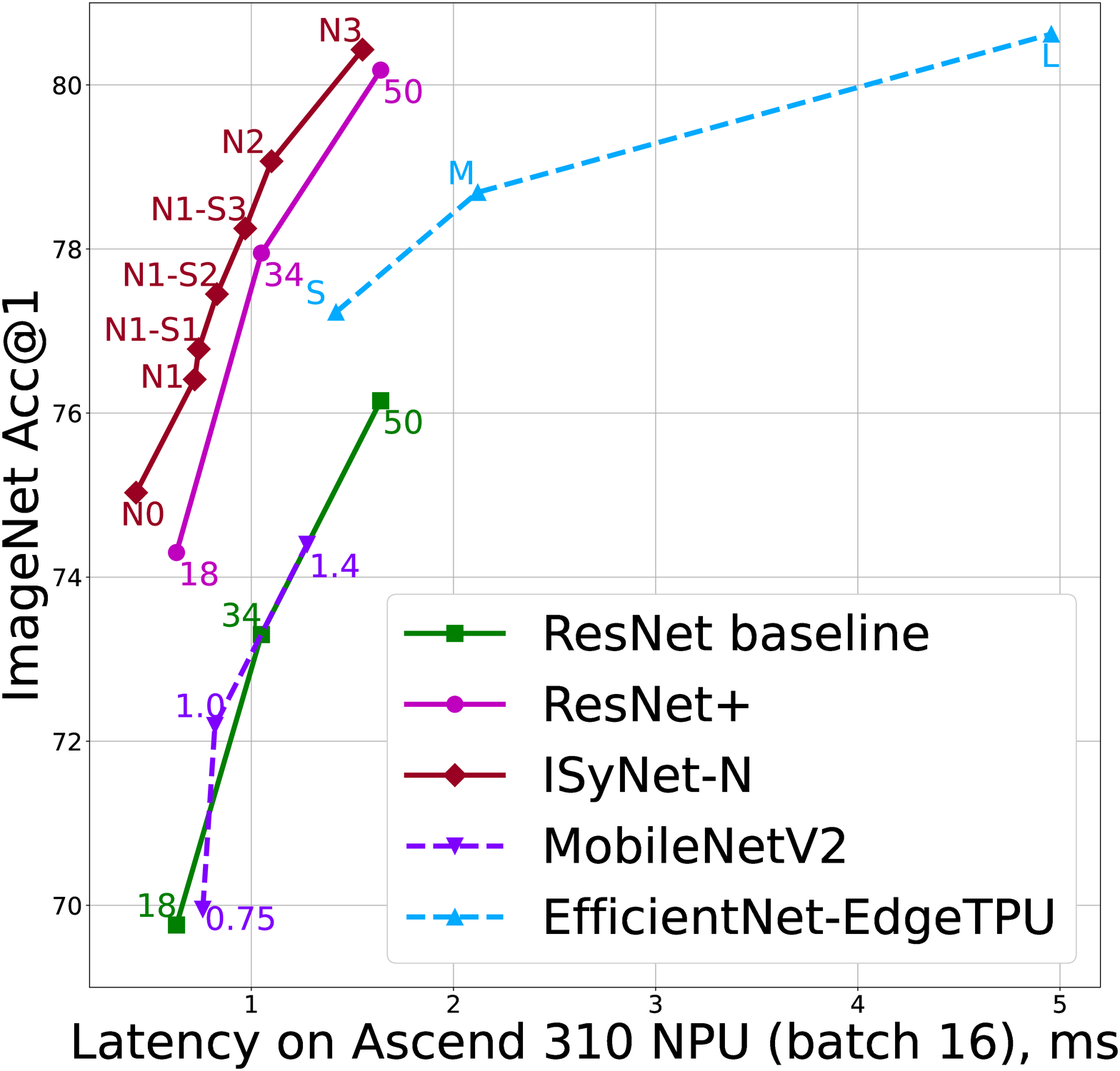}
\caption{Accuracy/latency trade-off for different models on ImageNet.
'ISyNet-N' curve shows our models results; 'ResNet baseline' denotes original ResNet results; 'ResNet+' denotes original ResNet models trained with our training procedure; 'MobileNetV2' and 'EfficientNet-EdgeTPU'~\cite{gupta2020acceleratoraware} curves denote original corresponding architectures.} \label{fig:isynet_imagenet}
\end{wrapfigure} 

\section{Background and Related Art} \label{background}
Search of accurate and fast architectures for computer vision was a tricky work for a long time. VGGNet~\cite{simonyan2015deep} with 19 convolutional layers was proposed in 2014 and showed the benefit of constructing complex powerful models. ResNet~\cite{he2015deep} architectures were proposed in 2015 to overcome the problem of training very deep networks. This work gave technology for successful training of architectures up to 200 layers with the help of residual connections.
One of the first and still popular research works considering model complexity is a MobileNet architectures family~\cite{howard2017mobilenets, sandler2019mobilenetv2}. Authors used depth-wise separable convolutions to build lightweight deep neural networks.

NAS was developed to change paradigm of neural networks construction and automate the work of  human researchers. NAS consists of 3 main parts: search space, search method and response function. The search space is defined as a subspace of all architectures limited by some manually defined constraints. The choice of the search space is a key point of NAS. However, it isn't deeply discussed in the literature. Typical search spaces are: hierarchical search space~\cite{tan2019mnasnet} and cell-based search space~\cite{ZophNasNet}. Main search  methods are reinforcement learning (RL)~\cite{zoph2016RL}, evolutionary algorithms (EA)~\cite{AmoebaReal}, surrogate-model based optimization (SMBO)~\cite{PNASNet,luo2019neural} and one-shot approaches~\cite{DARTS}. RL, EA and SMBO are time consuming, but found architectures outperform to those found by one-shot method~\cite{NASSurvey}. As a response function authors typically use task-specific metrics (classification top1 accuracy, detection mAP and so on) possibly combined with complexity penalty~\cite{ProxylessNAS}.

Thanks to NAS, automatically constructed architectures became state-of-the-art first for small datasets such as CIFAR-10~\cite{wang2021sampleefficient} and later for big datasets such as ImageNet~\cite{tan2020efficientnet}. 
Platform-aware NAS~\cite{tan2019mnasnet} allows to look for models with a good trade-off between the accuracy and the latency on the mobile devices. NAS requires a huge amount of computational resources, that's why it's computationally expensive to obtain architectures for all necessary complexity scenarios. To overcome this problem scaling methods are used ~\cite{tan2020efficientnet}. EfficientNet series of architectures obtained by NAS and scaled with compound scaling are now state-of-the-art on the ImageNet~\cite{pham2021meta}. While compound scaling considers FLOPS as a complexity measure of an architecture other researches show that total activations number in the model has stronger influence to the latency on the real hardware than the FLOPS~\cite{radosavovic2020designing}. Authors of the EfficientNet-EdgeTPU~\cite{gupta2020acceleratoraware} models family designed to be efficient on the Google Edge TPU devices conclude that theoretical computation measures (MACs, FLOPS) are not an optimal proxy for the real latency and use special cycle-accurate performance simulator.
In our work we use surrogate-model based optimization. This approach uses a surrogate model $\hat{f}$ to approximate the response function $f$. This model is trained on meta-dataset which contains architecture descriptors and their response values gathered during the architecture search: $H=\{(\alpha_1,~f(\alpha_1)),~(\alpha_2,~f(\alpha_2)),~...\}$. Generally, surrogate model is trained to minimize squared error: $\mathcal{L}=\sum_{\substack{(\alpha,~f(\alpha)) \in H}}(\hat{f}(\alpha)~-~f(\alpha))^2$. In practice very precise estimation is not necessary as long as surrogate provides a useful ranking to identify promising candidate. After candidate $\alpha^*$ is trained and evaluated corresponding meta instances $(\alpha^*,~f(\alpha^*)$ are added to $H$ and the surrogate model is updated.

\section{Problem} \label{problem}

\subsection{NPU design and constraints}
There exist a number of AI acceleration hardware (e.g. Google Cloud TPU~\cite{tpu2019edge, 10.1145/3154484}, NVidia Jetson~\cite{nvidiajetson}, Huawei Ascend~\cite{huawei2021ascend310}, Intel Movidius Myriad~\cite{intelmovidiusmyriad}), and these devices set a constraints for the DNNs to be deployed.
These devices are typically good at parallelizable tasks of tensor and matrix multiplications and additions as well as other operations commonly used in DNNs, such as activation functions and other element-wise operations.
In our paper we consider the optimization for Ascend 310 NPU based on DaVinci architecture~\cite{huawei2020davinci}, 
however, our model can be implemented for any AI accelerators that support 2D convolution with non-square kernels. Example of an application for this platform can be found in ~\cite{taskynov2021tensor}.

The AI core is a main part of Ascend 310 NPU and it executes tensor and vector operations.
Three main compute units of AI Core are: the cube unit, which performs the matrix multiplications, including the fully-connected layers and the convolutions;
the vector unit which executes the vector operations like an element-wise sum of tensors, batch normalization~\cite{batchnorm} and the activation functions;
the scalar unit which is responsible for the scalar operations and controls the program flow and addressing.

The cube unit performs multiplication of two 16x16 float16 matrices or 16x32 and 32x16 int8 matrices at a time, and this is one of the most important constraint imposed by the design of the cube unit.
Matrices of larger size are multiplied by parts.
If a size of multiplied matrices is less than specified they will be padded by zeros.
It is acceptable, but the highest cube unit utilization is reached for the matrices with a size divisible by 16 or 32 depending on the computation precision.
The vector unit is responsible for the vector computations.
It provides less computational power than the cube unit, but the capabilities of computations are more flexible.
To process and store the data there are several storage units in AI core and in some scenarios data data buses may become a bottleneck.

What properties should an NPU-efficient neural architecture search space have?
\begin{itemize}
\item For all shapes of tensors (including weights and activations) that are processed by the Cube Unit divisibility by 16 (considering float16 precision) is preferable.
\item Matrix operations are more preferable than vector operations and operations with data.
Operations that could be reduced to matrix multiplication (e.g. Convolutions and Fully-connected layers) are preferable over operations that can be done on the Vector Unit only.
The only exception is "operator fusion" mechanism that allows to fuse vector operations following matrix operation into one fused operation. An example of such efficient fusion is a sequence (Convolution~${\rightarrow}$~BatchNorm~${\rightarrow}$~Activation).

\item When possible it is better to avoid an element-wise product, sum and permutation.

\item Every operation in a computational graph requires input and output data to be transferred, which increases the total latency.
Branched architectures like DenseNet~\cite{huang2018densely} or Inception~\cite{szegedy2014going} should be avoided.
Chain-like graph is preferable. The number of residual connections should be minimized.
\item Lightweight activation functions (e.g. ReLU) are preferable over complex ones (e.g. Swish~\cite{ramachandran2017searching}, Mish~\cite{misra2020mish} or GELU~\cite{hendrycks2020gaussian})
\end{itemize}

\subsection{Matrix Efficiency Measure}

To evaluate efficiency of specific architectures and whole search spaces to NPU we propose a novel \textit{Matrix Efficiency Measure (MEM)}.
Let us consider three most important sources of latency during neural networks inference: matrix operations, vector operations and data transfer (including input, output and weights).
Scalar operations are negligible and not counted for simplicity.

For each operation $o_i$ in Neural Network $A$ we can define the following measures: $m(o_i)$ – the number of matrix operations; $v(o_i)$ – the number of vector operations. $d(o_i)$ – the number of input, output data and weights of the operation.
Matrix efficiency measure for architecture $A=\{o_1,o_2, \dots ,o_N\}$ can be estimated as follows:
\begin{equation}
\begin{aligned}
\boldsymbol{MEM}(A) = \dfrac{ w_m \cdot \sum_{i=1}^N m(o_i)} {\sum_{i=1}^N (w_m \cdot m(o_i) + w_v \cdot v(o_i) + w_d \cdot d(o_i))}
\end{aligned}
\end{equation}
$\forall A: \boldsymbol{MEM}(A) \in [0;1)$ and the closer $MEM(A)$ to 1 the more efficient $A$ to NPU design, because matrix operations are preferable over the others.
For a specific design space $D = \{A_1,A_2, \dots ,A_K \}$ mean matrix efficiency measure can be estimated as follows:
\begin{equation}
\begin{aligned}
\boldsymbol{mMEM}(D) = \dfrac{1}{K} \cdot \sum_{j=1}^K \boldsymbol{MEM}(A_j);
\end{aligned}
\end{equation}
The closer $mMEM(D)$ to 1 the more efficient whole design space $D$ to NPU design.

To find values of $w_m$, $w_v$ and $w_d$ we train the following linear regression model that approximates latency of architecture $A$:
\begin{equation}
\begin{aligned}
\boldsymbol{lat}(A) = w_0 + w_m \cdot \sum_{i=1}^N m(o_i) + w_v \cdot \sum_{i=1}^N v(o_i) + w_d \cdot \sum_{i=1}^N d(o_i)
\end{aligned}
\end{equation}
In our experiments we found that: $w_0 = 0.773$; $w_m = 2.57\mathrm{e}{-9}$; $w_v = -1.26\mathrm{e}{-8}$; $w_d = 3.36\mathrm{e}{-8}$. This model has mean absolute percentage error $12.54\%$ and coefficient of determination $0.94$ which is reasonable for such a simple model. Negative value of $w_v$ can be explained by the fact that $m(o_i)$, $v(o_i)$ and $d(o_i)$ are not linearly independent.
We used linear regression model because of its simplicity and interpretability.
Ablation study for the latency models is presented in the supplementary materials.

\begin{figure}[t]
\begin{center}
\includegraphics[width=1.0\linewidth]{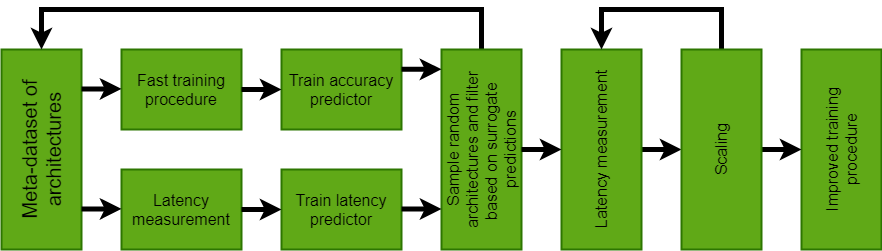}
\end{center}
  \caption{Overall scheme of ISyNAS - NPU-aware architectures search algorithm. Algorithm is based on 2 surrogate models (SM). One for accuracy and another for latency. After gathering meta-dataset of architectures and training SMs we sample architectures with limited latency and good estimation of accuracy. Meta-dataset updated and SMs retrained. Best found models scaled with scaling procedure and trained with long improved training procedure.}
\label{fig:isynas_scheme}
\end{figure}

\begin{table}[b!]
\centering
\small
\begin{tabular}{|c|c|c|c|c|}
\hline
Operation & \#Matrix ops. & \#Vector ops. & Data ops. & MEM(op) \cr
\hline\hline
$Conv7x7$ & $2.9\mathrm{e}{+11}$ & 0 & $2.9\mathrm{e}{+7}$ & $9.7\mathrm{e}{-1}$ \cr
\hline
$Conv5x5$ & $1.5\mathrm{e}{+11}$ & 0 & $2.0\mathrm{e}{+7}$ & $9.6\mathrm{e}{-1}$ \cr
\hline
$Conv3x3$ & $5.4\mathrm{e}{+10}$ & 0 & $1.4\mathrm{e}{+7}$ & $9.2\mathrm{e}{-1}$ \cr
\hline
$Conv7x1$ & $4.2\mathrm{e}{+10}$ & 0 & $1.3\mathrm{e}{+7}$ & $9.0\mathrm{e}{-1}$ \cr
\hline
$Conv5x1$ & $2.3\mathrm{e}{+10}$ & 0 & $1.2\mathrm{e}{+7}$ & $8.8\mathrm{e}{-1}$ \cr
\hline
$Conv3x1$ & $1.8\mathrm{e}{+10}$ & 0 & $1.1\mathrm{e}{+7}$ & $8.2\mathrm{e}{-1}$ \cr
\hline
$Conv1x1$ & $6.0\mathrm{e}{+9}$ & 0 & $1.05\mathrm{e}{+7}$ & $6.2\mathrm{e}{-1}$ \cr
\hline
$DepthWiseConv5x5$ & $3.3\mathrm{e}{+8}$ & 0 & $1.01\mathrm{e}{+7}$ & $8.6\mathrm{e}{-2}$ \cr
\hline
$DepthWiseConv3x3$ & $1.2\mathrm{e}{+8}$ & 0 & $1.01\mathrm{e}{+7}$ & $3.2\mathrm{e}{-2}$ \cr
\hline
$Pooling3x3$ & 0 & $2.2\mathrm{e}{+7}$ & $1.1\mathrm{e}{+7}$ & 0 \cr
\hline
$ReLU$ & 0 & $6.5\mathrm{e}{+6}$ & $1.3\mathrm{e}{+7}$ & 0 \cr
\hline
$ReLU6$ & 0 & $1.3\mathrm{e}{+7}$ & $1.3\mathrm{e}{+7}$ & 0 \cr
\hline
$Swish$ & 0 & $1.3\mathrm{e}{+7}$ & $1.3\mathrm{e}{+7}$ & 0 \cr
\hline
$BatchNorm$ & 0 & $5.2\mathrm{e}{+7}$ & $1.3\mathrm{e}{+7}$ & 0 \cr
\hline
$ElementwiseAdd$ & 0 & $6.5\mathrm{e}{+6}$ & $1.9\mathrm{e}{+7}$ & 0 \cr
\hline
$Concatenation$ & 0 & 0 & $2.6\mathrm{e}{+7}$ & 0 \cr
\hline
\end{tabular}
\caption{Averaged matrix, vector, data and $MEM$ characteristics of different operations}
\label{table:mem_ops}
\end{table}

Table~\ref{table:mem_ops} shows hardware efficiency of operations according to our model.
$Conv7x7$, $Conv5x5$, $Conv3x3$, $Conv1x1$ and convolutions with non-squared kernel (7x1, 5x1, 3x1) show similar efficiency.
Depthwise convolutions efficiency are less by one order. Operations which do not have matrix operations (poolings, activations, BatchNorms, addition and concatenation) are considered as non-efficient for NPU.

\begin{wraptable}[12]{r}{60mm}
    \begin{tabular}{|c|c|c|}
    \hline
    Design space & \#arch & $mMEM$  \cr
    \hline\hline
    ResNet & 1000 & 0.294 \cr
    \hline
    MobileNetV2 & 100 & 0.167 \cr
    \hline
    MNasNet & 100 & 0.17 \cr
    \hline
    ISyNet-N (ours) & 3800 & 0.378 \cr
    \hline
    \end{tabular}
\caption{$mMEM$ of different design spaces. Average $mMEM$ of proposed design space is better, then ResNet due to flexible block length and skip-connections. MobileNet and MNasNet search space $mMEM$ is worse due to non-optimal NPU operations. }
\label{table:mmem_spaces}
\end{wraptable}

However, BatchNorm and activation placed after convolution operation can be efficiently fused into one operator and do not result in significant slowdown at the inference stage.
Elementwise addition and concatenation operations are widely used in residual blocks, which is essential for convergence of the model training.
Thus, we can't avoid these blocks at all, but can use them flexibly, depending on real impact to the model properties.

\section{Method}
\subsection{Search space design}\label{section:search_space}

To optimize architecture search we propose the search space, which incorporates only suitable for NPU operations and use flexible block length, width to add additional freedom in NAS.
This search space is defined by the following set of rules:

\begin{itemize}
\item Model is divided into $NS$ stages, $NS\in[1,..,6]$ is not fixed, but limited;
\item Every stage $s\in[1,...,NS]$ is divided into blocks $B_s$ of identical structure. The number of blocks $NB_s\in[1,..,20]$ in each stage is not fixed, but limited;
\item Every block $B_s$ is divided into four edges ($E_{s,i}$) each of them can be a convolution with kernel 1x1, 3x3, 5x5, 7x7, sequence of convolutions with non-square kernels 1x3+3x1, 1x5+5x1, 1x7+7x1 or an identity operation;
\item Output of the block could be summed up with its input by a skip connection which is defined by the flag $SK_s\in[0,1]$ for all blocks in the stage $s$;
\item Each convolution edge has the follow-up normalization operation (BatchNorm) and coupled with an activation (ReLU)
\item Last non-identity edge could be uncoupled with activation which is defined by flag $LA_s\in[0,1]$ for all blocks in the stage $s$.
\item The number of output channels of block $B_i$ is $2^{3+i+CI_s}$, where $CI_s\in[0,2]$ is searched non-negative integer defined for all blocks in the stage $s$;
\item The number of output channels for every intermediate (except the last one) edge of blocks in the stage $s$ is multiplied by $EF_s$ – positive integer searched parameter;
\item First block of stage differs from all others. It has stride 2 in the first non-identity edge and no skip connection
\end{itemize}

Flexible block length and skip connections together with hardware-efficient operations make the proposed search space better for hardware-targeted NAS.
Comparison of mMEM values for different search spaces are shown in the table~\ref{table:mmem_spaces}.
ISyNet space outperforms ResNet~\cite{he2015deep}, MobileNetV2~\cite{sandler2019mobilenetv2} and MNasNet~\cite{tan2019mnasnet} search spaces.

\subsection{Search method}

As it is shown in ~\cite{radosavovic2020designing,gupta2020acceleratoraware} FLOPS and MACs are not the best option to measure the complexity for the real hardware. In our work we use the accuracy and the latency surrogate models for efficient and more accurate search of the architectures and scaling. Overall scheme of ISyNAS, NPU-aware neural architecture search approach, is shown on the Figure~\ref{fig:isynas_scheme}. We use a surrogate model (SM) based optimization. To train SM we encode each architecture according to the proposed notation of search space design with following vector:

\begin{equation}
\begin{gathered}
E = ( NS, [V_1, V_2, ..., V_{NS}]), \\
V_s = (LA_s,\;NB_s,\;EF_s,\;SK_s,\;CI_s,\;E_{s,0},\;E_{s,1},\;E_{s,2},\;E_{s,3}), s \in [1,\;...,\;NS]
\end{gathered}
\end{equation}

The value of $E_{s,i}\in[0,7]$ is coded with following encoding: (conv1x1: 0, conv3x3: 1, conv5x5: 2, conv7x7: 3, conv1x3+conv3x1: 4, conv1x5+conv5x1: 5, conv1x7+conv7x1: 6, Identity: 7). If $NS<6$ encoding vector is padded with zeros. For SM we use LSTM architecture to process the encoded vector and apply fully-connected layer to the resulting embedding. We train two different SM for the accuracy and the latency. To train SM we collected dataset of 400 architectures trained on ImageNet and measured on Ascend 310 NPU with batch16.

\begin{wrapfigure}[22]{r}{70mm}
\includegraphics[width=1.0\linewidth]{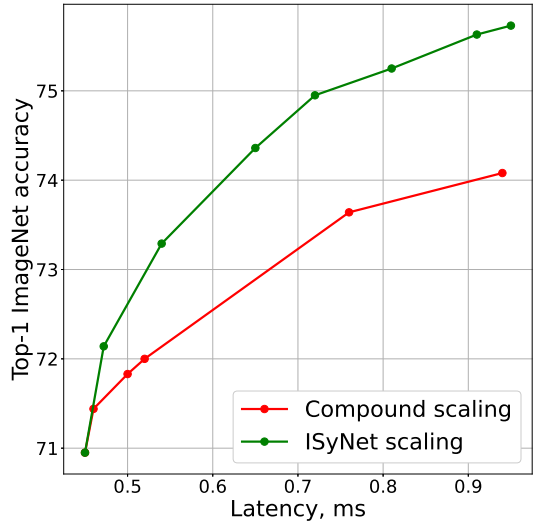}
\caption{Accuracy/latency trade-off for our ISyNet scaling vs compound scaling. ISyNet scaling shows better results than the compound scaling due to heterogeneous stages scaling and targeting to the NPU latency.}
\label{fig:scaling}
\end{wrapfigure}

We use following training procedure to collect dataset: SGD optimizer with momentum 0.9, 2 epochs warmup to maximum learning rate 0.4, 120 epochs with exponential learning rate decay, multiplying by 0.1 every 30 epochs, weight decay 0.0001, 8 NVidia V100 GPUs, total batch size 1024, O1 optimization.

\subsection{Scaling algorithm}

Tan et al.~\cite{tan2020efficientnet} proposed a method of compound architecture scaling which performs the small exhaustive search of the scaling parameters for width, depth and input resolution. For this purpose we select the parameters of architecture related to depth ($NB_s$) and width ($EF_s$, $CI_s$) and measure their impact on the accuracy and the latency. We have empirically found that increase of parameters related to the depth provide better accuracy/latency trade-off, than increase of parameters related to width. Out result matches experimental results obtained in ~\cite{FastModelsFamilies}.
Scaling of the input resolution affects only ImageNet performance, while does not affect the downstream tasks which are often use different resolution.
Thus, we scale only depth of models and use the different scaling coefficients for the different stages.
We use the latency as a scaling complexity function by carefully estimating latency of blocks.
We perform the small brute-force search of scaling coefficients to find Pareto frontier of the scaled architectures.
The comparison of our scaling method with the compound scaling is shown on the figure~\ref{fig:scaling}.

\section{Experiments}
\subsection{Architectures}
Found ISyNet architectures are described in the table~\ref{table:architectures}.
Every architecture is defined by its stages.
Stages are defined by block patterns and number of blocks ($NB$).
Block patterns are defined by the list of operations, presence of last activation ($LA$) and skip connection ($SK$).
Architectures N0, N1, N2, N3 are found by NAS, while N1-S1, N1-S2, N1-S3 are found by scaling method from architecture $N1$. The process of the search took about 20000 GPU*hours which is comparable with the other approaches~\cite{NASSurvey}.

\subsection{ImageNet experiments}

To train our models on ImageNet~\cite{ILSVRC15} we use one server with 8 Ascend 910 NPUs and the following training setup: AdamW optimizer; 40 epochs warmup to maximum learning rate 0.001; 550 epochs with cosine learning rate decay,  weight decay 0.05; batch size 1024 (128 per device); RandAugment~\cite{cubuk2019randaugment} augmentation policy; Deep Mutual Learning~\cite{zhang2017deep} with larger model (for ISyNet models ISyNet-N3, for ResNet models ResNet-101), BatchNorm after the last fully-connected layer ("LastBN"), label smoothing.
The ablation study for the listed tricks is presented in the supplementary materials.
Training results are shown in the table~\ref{table:isynet_imagenet_table} and Pareto frontiers are illustrated on the figure~\ref{fig:isynet_imagenet}. For the fair comparison we train the ResNet models with our training procedure and they are presented in the table and figure as ResNet+. For the MobileNetV2 architectures our procedure does not bring an improvement and we use an original results.
Latency of all resulting architectures are measured on the Ascend 310 NPU with batch size 16 and reduced to a single image.

\begin{table*}[h!]
\centering
\small
\begin{tabular}{|c|c|c|p{1cm}|c|c|}
\hline
model & stage & operations in block & $LA$ & $NB$ & $SK$ \cr
\hline
\multirow{5}{*}{N0} & 0 & \small{5x5x16} & 1 & 1 & 0 \cr
\cline{2-6}
& 1 & \small{3x3x32, 3x3x32} & 1 & 2 & 1 \cr
\cline{2-6}
& 2 & \small{3x3x64, 3x3x64, 1x1x64, 1x1x64} & 1 & 4 & 1 \cr
\cline{2-6}
& 3 & \small{1x1x128, 3x3x128} & 0 & 2 & 1 \cr
\cline{2-6}
& 4 & \small{3x3x256, 3x3x256, 1x1x256} & 1 & 6 & 1 \cr
\hline
\multirow{5}{*}{N1} & 0 & \small{7x7x16} & 1 & 1 & 1 \cr
\cline{2-6}
& 1 & \small{1x5x32+5x1x32, 1x1x32, 3x3x32, 5x5x32} & 1 & 1 & 1 \cr
\cline{2-6}
& 2 & \small{3x3x128, 3x3x128, 3x3x128} & 1 & 4 & 1 \cr
\cline{2-6}
& 3 & \small{3x3x128, 1x1x128, 3x3x128} & 1 & 6 & 1 \cr
\cline{2-6}
& 4 & \small{1x1x256, 1x1x256, 1x1x256} & 1 & 1 & 1 \cr
\hline
\multirow{5}{*}{N1-S1} & 0 & \small{7x7x16} & 1 & 1 & 1 \cr
\cline{2-6}
& 1 & \small{1x5x32+5x1x32, 1x1x32, 3x3x32, 5x5x32} & 1 & 1 & 1 \cr
\cline{2-6}
& 2 & \small{x3x128, 3x3x128, 3x3x128} & 1 & 4 & 1 \cr
\cline{2-6}
& 3 & \small{3x3x128, 1x1x128, 3x3x128} & 1 & 6 & 1 \cr
\cline{2-6}
& 4 & \small{1x1x256, 1x1x256, 1x1x256} & 1 & 3 & 1 \cr
\hline
\multirow{5}{*}{N1-S2} & 0 & \small{7x7x16} & 1 & 1 & 1 \cr
\cline{2-6}
& 1 & \small{1x5x32+5x1x32, 1x1x32, 3x3x32, 5x5x32} & 1 & 1 & 1 \cr
\cline{2-6}
& 2 & \small{3x3x128, 3x3x128, 3x3x128} & 1 & 5 & 1 \cr
\cline{2-6}
& 3 & \small{3x3x128, 1x1x128, 3x3x128} & 1 & 6 & 1 \cr
\cline{2-6}
& 4 & \small{1x1x256, 1x1x256, 1x1x256} & 1 & 6 & 1 \cr
\hline
\multirow{5}{*}{N1-S3} & 0 & \small{7x7x16} & 1 & 1 & 1\cr
\cline{2-6}
& 1 & \small{1x5x32+5x1x32, 1x1x32, 3x3x32, 5x5x32} & 1 & 1 & 1 \cr
\cline{2-6}
& 2 & \small{3x3x128, 3x3x128, 3x3x128} & 1 & 6 & 1 \cr
\cline{2-6}
& 3 & \small{3x3x128, 1x1x128, 3x3x128} & 1 & 8 & 1 \cr
\cline{2-6}
& 4 & \small{1x1x256, 1x1x256, 1x1x256} & 1 & 7 & 1 \cr
\hline
\multirow{5}{*}{N2} & 0 & \small{3x3x16, 7x7x16,7x7x16} & 1 & 1 & 1 \cr
\cline{2-6}
& 1 & \small{5x5x64, 3x3x64, 3x3x32} & 0 & 3 & 1 \cr
\cline{2-6}
& 2 & \small{3x3x64, 3x3x64, 3x3x64} & 1 & 4 & 1 \cr
\cline{2-6}
& 3 & \small{3x3x128, 1x1x128} & 0 & 17 & 1 \cr
\cline{2-6}
& 4 & \small{1x1x512, 1x1x512, 1x1x512} & 1 & 2 & 1 \cr
\hline
\multirow{5}{*}{N3} & 0 & \small{5x5x32, 7x7x32} & 1 & 1 & 0 \cr
\cline{2-6}
& 1 & \small{3x3x64, 3x3x64} & 0 & 5 & 1 \cr
\cline{2-6}
& 2 & \small{3x3x128,1x3x128+3x1x128,3x3x128, 3x3x128} & 1 & 3 & 1 \cr
\cline{2-6}
& 3 & \small{1x3x256+3x1x256, 1x1x256} & 0 & 13 & 1 \cr
\cline{2-6}
& 4 & \small{1x1x1024, 1x1x1024, 1x1x1024} & 1 & 1 & 1 \cr
\hline
\end{tabular}
\caption{Specification of found architectures.}
\label{table:architectures}
\end{table*}

\begin{table*}[h!]
\centering
\small
\begin{tabular}{|c|c|c|c|c|c|}
\hline
Model & Top-1 acc. & Latency, ms. & \# Params, $\times 10^6$ & MACs, $\times 10^9$ & MEM \cr
\hline\hline
ISyNet-N0 & 75.03 & 0.43 & 9.59 & 1.13 & 0.214\cr 
\hline
ResNet-18+ & 74.3 & 0.63 & 11.69 & 2.28 & 0.439\cr
\hline
ISyNet-N1 & 76.41 & 0.72 & 7.42 & 2.85 & 0.399\cr 
\hline
ISyNet-N1-S1 & 76.78 & 0.74 & 7.82 & 2.88 & 0.391\cr 
\hline
ISyNet-N1-S2 & 77.45 & 0.83 & 8.86 & 3.34 & 0.395\cr 
\hline
ISyNet-N1-S3 & 78.25 & 0.97 & 10.81 & 4.12 & 0.399\cr 
\hline
ResNet-34+ & 77.95 & 1.05 & 21.8 & 4.63 & 0.497\cr
\hline
ISyNet-N2 & 79.07 & 1.1 & 19.43 & 4.93 & 0.351\cr 
\hline
ISyNet-N3 & 80.43 & 1.55 & 20.47 & 7.32 & 0.394\cr 
\hline
ResNet-50+ & 80.18 & 1.64 & 25.56 & 5.19 & 0.286\cr
\hline
\end{tabular}
\caption{ISyNet performance results on ImageNet. Table shows, that ResNet-18 and ResNet-34 have very good MEM due to optimization of NPU devices to these specific architectures. In the same time average MEM of ResNet space has lower MEM according to table ~\ref{table:mmem_spaces}. It allow us to find architectures in our space, which outperform ResNets. We don't show results for other type of architectures like MobileNet, EfficientNet as they are non-efficient on NPU according to figure ~\ref{fig:isynet_imagenet}.}
\label{table:isynet_imagenet_table}
\end{table*}

\subsection{Transfer learning experiments}
\begin{figure*}[t]
\begin{center}
\includegraphics[width=0.9\linewidth]{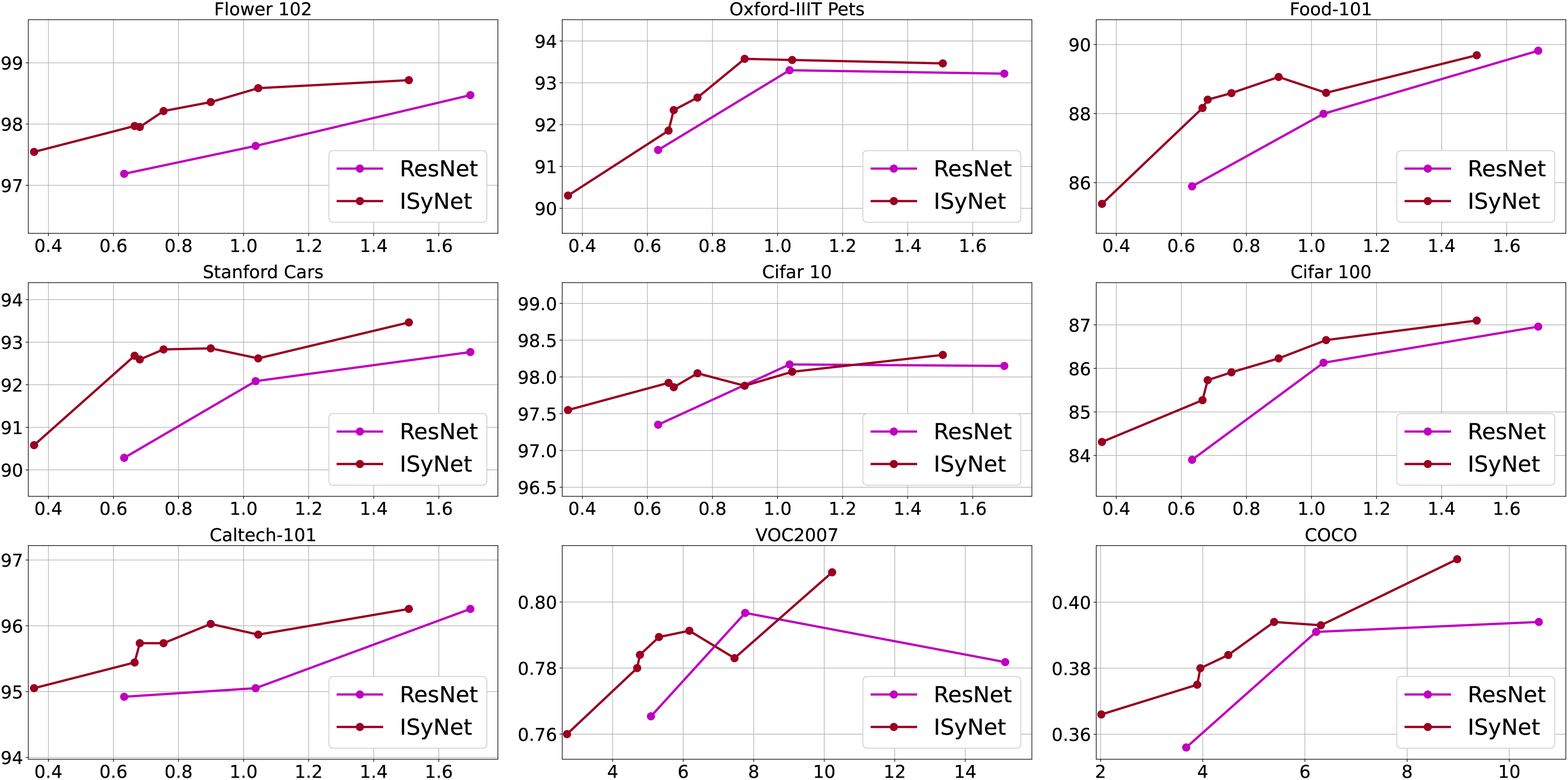}
\end{center}
   \caption{Results of transfer learning for different downstream tasks. Latency (in ms.) is plotted along the horizontal axis. Target metrics on the corresponding dataset are plotted along the vertical axis of each plot. Curves show stable improvement over ResNet architectures both for classification downstream datasets and as backbone for detection datasets.}
\label{fig:isynet_transfer}
\end{figure*}

To verify generalization ability of ISyNet models we do the transfer learning experiments for the following downstream datasets and tasks: classification (CIFAR-10, CIFAR-100~\cite{Krizhevsky09learningmultiple}, Caltech 101~\cite{1597116}, Flowers~\cite{Nilsback08}, Oxford-IIIT Pet~\cite{parkhi12a}, Stanford Cars~\cite{KrauseStarkDengFei-Fei_3DRR2013}, Food-101~\cite{bossard14}) and object detection (Pascal VOC 2007~\cite{pascalvoc}, as a backbone for YOLOv3~\cite{redmon2018yolov3}; MS COCO~\cite{lin2015microsoft}, as a backbone for Faster R-CNN~\cite{ren2016faster}). The details about experimental setup are presented in the supplementary materials.
Results of transfer learning are presented in the Table~\ref{table:isynet_transfer_table} and illustrated on the Figure~\ref{fig:isynet_transfer}. We compare our models with the results of ResNet+ architectures trained with the same procedure as ISyNet. In most cases our architectures outperform results of ResNet which proves good generalization ability of the ISyNet.

\begin{table*}[h!]
\centering
\small
\begin{tabular}{|c|c|c|c|c|c|c|c|c|c|}
\hline
Model & \rotatebox[origin=c]{90}{CIFAR-10} & \rotatebox[origin=c]{90}{CIFAR-100} & \rotatebox[origin=c]{90}{Caltech-101} & \rotatebox[origin=c]{90}{Flowers} & \rotatebox[origin=c]{90}{Oxford-IIIT Pet} & \rotatebox[origin=c]{90}{Stanford Cars} & \rotatebox[origin=c]{90}{Food-101} & \rotatebox[origin=c]{90}{VOC 2007} & \rotatebox[origin=c]{90}{COCO} \cr
\hline\hline
ISyNet-N0 & 97.55 & 84.31 & 95.05 & 97.55 & 90.31 & 90.58 & 85.39 & 76 & 36.6 \cr 
\hline
ResNet-18+ & 97.35 & 83.9 & 94.92 & 97.19 & 91.39 & 90.28 & 85.89 & 76.54 & 35.6 \cr
\hline
ISyNet-N1 & 97.92 & 85.27 & 95.44 & 97.97 & 91.86 & 92.68 & 88.16 & 78 & 37.5 \cr 
\hline
ISyNet-N1-S1 & 97.86 & 85.73 & 95.74 & 97.95 & 92.35 & 92.59 & 88.40 & 78.40 & 38.0 \cr 
\hline
ISyNet-N1-S2 & 98.05 & 85.91 & 95.74 & 98.21 & 92.65 & 92.83 & 88.59 & 78.94 & 38.4 \cr 
\hline
ISyNet-N1-S3 & 97.88 & 86.23 & 96.03 & 98.36 & 93.57 & 92.86 & 89.06 & 79.13 & 39.4 \cr 
\hline
ResNet-34+ & 98.17 & 86.13 & 95.05 & 97.64 & 93.30 & 92.09 & 87.99 & 79.67 & 39.1 \cr
\hline
ISyNet-N2 & 98.07 & 86.65 & 95.87 & 98.59 & 93.55 & 92.62 & 88.60 & 78.30 & 39.3 \cr 
\hline
ISyNet-N3 & 98.3 & 87.1 & 96.26 & 98.72 & 93.46 & 93.46 & 89.69 & 80.90 & 41.3 \cr 
\hline
ResNet-50+ & 98.15 & 86.96 & 96.26 & 98.47 & 93.22 & 92.77 & 89.82 & 78.18 & 39.4 \cr
\hline
\end{tabular}
\caption{ISyNet transfer learning performance results on the downstream tasks}
\label{table:isynet_transfer_table}
\end{table*}

\section{Conclusions}

In this paper we study the problem of efficiency of neural networks for NPU devices - specialized AI accelerators.
To address the question of ``NPU-efficiency`` estimation we propose $MEM$ - novel measure of matrix computations efficiency in Neural Networks.
With the help of $MEM$ we design the search space for our convolutional backbones and do Neural Architecture Search in this space.
Finally, we propose ISyNet - the family of NPU-efficient convolutional backbones that outperform strong NPU-efficient baselines by a significant margin and prove good generalization properties of ISyNet on many Computer Vision datasets and tasks.

\clearpage
\bibliographystyle{ieee_fullname.bst}
\bibliography{egbib}

\begin{thebibliography}{10}\itemsep=-1pt

\bibitem{bossard14}
Lukas Bossard, Matthieu Guillaumin, and Luc Van~Gool.
\newblock Food-101 -- mining discriminative components with random forests.
\newblock In {\em European Conference on Computer Vision}, 2014.

\bibitem{ProxylessNAS}
Han Cai, Ligeng Zhu, and Song Han.
\newblock Proxylessnas: Direct neural architecture search on target task and
  hardware, 2018.

\bibitem{cubuk2019randaugment}
Ekin~D. Cubuk, Barret Zoph, Jonathon Shlens, and Quoc~V. Le.
\newblock Randaugment: Practical automated data augmentation with a reduced
  search space, 2019.

\bibitem{pascalvoc}
M. Everingham, L. Van~Gool, and C.K.I. et~al. Williams.
\newblock The pascal visual object classes (voc) challenge.
\newblock {\em Int J Comput Vis}, 88:303–338, 2010.

\bibitem{tpu2019edge}
Google.
\newblock {Edge TPU}, 2019.

\bibitem{gupta2020acceleratoraware}
Suyog Gupta and Berkin Akin.
\newblock Accelerator-aware neural network design using automl, 2020.

\bibitem{he2015deep}
Kaiming He, Xiangyu Zhang, Shaoqing Ren, and Jian Sun.
\newblock Deep residual learning for image recognition, 2015.

\bibitem{hendrycks2020gaussian}
Dan Hendrycks and Kevin Gimpel.
\newblock Gaussian error linear units (gelus), 2020.

\bibitem{howard2017mobilenets}
Andrew~G. Howard, Menglong Zhu, Bo Chen, Dmitry Kalenichenko, Weijun Wang,
  Tobias Weyand, Marco Andreetto, and Hartwig Adam.
\newblock Mobilenets: Efficient convolutional neural networks for mobile vision
  applications, 2017.

\bibitem{huang2018densely}
Gao Huang, Zhuang Liu, Laurens van~der Maaten, and Kilian~Q. Weinberger.
\newblock Densely connected convolutional networks, 2018.

\bibitem{huawei2020davinci}
Huawei.
\newblock {Da Vinci Architecture}, 2020.

\bibitem{huawei2021ascend310}
Huawei.
\newblock {Ascend 310 AI Processor}, 2021.

\bibitem{intelmovidiusmyriad}
Intel.
\newblock {Intel Movidius Myriad™ X Vision Processing Unit}, 2021.

\bibitem{batchnorm}
Sergey Ioffe and Christian Szegedy.
\newblock Batch normalization: Accelerating deep network training by reducing
  internal covariate shift.
\newblock In {\em Proceedings of the 32nd International Conference on Machine
  Learning (ICML)}, 2015.

\bibitem{izmailov2019averaging}
Pavel Izmailov, Dmitrii Podoprikhin, Timur Garipov, Dmitry Vetrov, and
  Andrew~Gordon Wilson.
\newblock Averaging weights leads to wider optima and better generalization,
  2019.

\bibitem{10.1145/3154484}
Norman~P. Jouppi, Cliff Young, Nishant Patil, and David Patterson.
\newblock A domain-specific architecture for deep neural networks.
\newblock {\em Commun. ACM}, 61(9):50–59, Aug. 2018.

\bibitem{KrauseStarkDengFei-Fei_3DRR2013}
Jonathan Krause, Michael Stark, Jia Deng, and Li Fei-Fei.
\newblock 3d object representations for fine-grained categorization.
\newblock In {\em 4th International IEEE Workshop on 3D Representation and
  Recognition (3dRR-13)}, Sydney, Australia, 2013.

\bibitem{Krizhevsky09learningmultiple}
Alex Krizhevsky.
\newblock Learning multiple layers of features from tiny images.
\newblock Technical report, 2009.

\bibitem{FastModelsFamilies}
Sheng Li, Mingxing Tan, Ruoming Pang, Andrew Li, Liqun Cheng, Quoc Le, and
  Norman~P. Jouppi.
\newblock Searching for fast model families on datacenter accelerators, 2021.

\bibitem{1597116}
{Li Fei-Fei}, R. {Fergus}, and P. {Perona}.
\newblock One-shot learning of object categories.
\newblock {\em IEEE Transactions on Pattern Analysis and Machine Intelligence},
  28(4):594--611, 2006.

\bibitem{lin2015microsoft}
Tsung-Yi Lin, Michael Maire, Serge Belongie, Lubomir Bourdev, Ross Girshick,
  James Hays, Pietro Perona, Deva Ramanan, C.~Lawrence Zitnick, and Piotr
  Dollár.
\newblock Microsoft coco: Common objects in context, 2015.

\bibitem{PNASNet}
Chenxi Liu, Barret Zoph, Maxim Neumann, Jonathon Shlens, Li-Jia~Li Wei~Hua, Li
  Fei-Fei, Alan Yuille, Jonathan Huang, and Kevin Murphy.
\newblock Progressive neural architecture search, 2017.

\bibitem{DARTS}
Hanxiao Liu, Karen Simonyan, and Yiming Yang.
\newblock Darts: Differentiable architecture search, 2018.

\bibitem{https://doi.org/10.48550/arxiv.1711.05101}
Ilya Loshchilov and Frank Hutter.
\newblock Decoupled weight decay regularization, 2017.

\bibitem{luo2019neural}
Renqian Luo, Fei Tian, Tao Qin, Enhong Chen, and Tie-Yan Liu.
\newblock Neural architecture optimization, 2019.

\bibitem{mallat_omp}
S.G. Mallat and Zhifeng Zhang.
\newblock Matching pursuits with time-frequency dictionaries.
\newblock {\em IEEE Transactions on Signal Processing}, 41(12):3397--3415,
  1993.

\bibitem{misra2020mish}
Diganta Misra.
\newblock Mish: A self regularized non-monotonic activation function, 2020.

\bibitem{Nilsback08}
Maria-Elena Nilsback and Andrew Zisserman.
\newblock Automated flower classification over a large number of classes.
\newblock In {\em Indian Conference on Computer Vision, Graphics and Image
  Processing}, Dec 2008.

\bibitem{nvidiajetson}
NVidia.
\newblock {Embedded Systems for Next-Generation Autonomous Machines}, 2021.

\bibitem{parkhi12a}
Omkar~M. Parkhi, Andrea Vedaldi, Andrew Zisserman, and C.~V. Jawahar.
\newblock Cats and dogs.
\newblock In {\em IEEE Conference on Computer Vision and Pattern Recognition},
  2012.

\bibitem{pham2021meta}
Hieu Pham, Zihang Dai, Qizhe Xie, Minh-Thang Luong, and Quoc~V. Le.
\newblock Meta pseudo labels, 2021.

\bibitem{radosavovic2020designing}
Ilija Radosavovic, Raj~Prateek Kosaraju, Ross Girshick, Kaiming He, and Piotr
  Dollár.
\newblock Designing network design spaces, 2020.

\bibitem{ramachandran2017searching}
Prajit Ramachandran, Barret Zoph, and Quoc~V. Le.
\newblock Searching for activation functions, 2017.

\bibitem{AmoebaReal}
Esteban Real, Alok Aggarwal, Yanping Huang, and Quoc~V Le.
\newblock Regularized evolution for image classifier architecture search, 2018.

\bibitem{redmon2018yolov3}
Joseph Redmon and Ali Farhadi.
\newblock Yolov3: An incremental improvement, 2018.

\bibitem{NASSurvey}
Pengzhen Ren, Yun Xiao, Xiaojun Chang, Po-Yao Huang, Zhihui Li, Xiaojiang Chen,
  and Xin Wang.
\newblock A comprehensive survey of neural architecture search: Challenges and
  solutions, 2020.

\bibitem{ren2016faster}
Shaoqing Ren, Kaiming He, Ross Girshick, and Jian Sun.
\newblock Faster r-cnn: Towards real-time object detection with region proposal
  networks, 2016.

\bibitem{ILSVRC15}
Olga Russakovsky, Jia Deng, Hao Su, Jonathan Krause, Sanjeev Satheesh, Sean Ma,
  Zhiheng Huang, Andrej Karpathy, Aditya Khosla, Michael Bernstein,
  Alexander~C. Berg, and Li Fei-Fei.
\newblock {ImageNet Large Scale Visual Recognition Challenge}.
\newblock {\em International Journal of Computer Vision (IJCV)},
  115(3):211--252, 2015.

\bibitem{sandler2019mobilenetv2}
Mark Sandler, Andrew Howard, Menglong Zhu, Andrey Zhmoginov, and Liang-Chieh
  Chen.
\newblock Mobilenetv2: Inverted residuals and linear bottlenecks, 2019.

\bibitem{simonyan2015deep}
Karen Simonyan and Andrew Zisserman.
\newblock Very deep convolutional networks for large-scale image recognition,
  2015.

\bibitem{szegedy2014going}
Christian Szegedy, Wei Liu, Yangqing Jia, Pierre Sermanet, Scott Reed, Dragomir
  Anguelov, Dumitru Erhan, Vincent Vanhoucke, and Andrew Rabinovich.
\newblock Going deeper with convolutions, 2014.

\bibitem{7780677}
Christian Szegedy, Vincent Vanhoucke, Sergey Ioffe, Jon Shlens, and Zbigniew
  Wojna.
\newblock Rethinking the inception architecture for computer vision.
\newblock In {\em 2016 IEEE Conference on Computer Vision and Pattern
  Recognition (CVPR)}, pages 2818--2826, 2016.

\bibitem{tan2019mnasnet}
Mingxing Tan, Bo Chen, Ruoming Pang, Vijay Vasudevan, Mark Sandler, Andrew
  Howard, and Quoc~V. Le.
\newblock Mnasnet: Platform-aware neural architecture search for mobile, 2019.

\bibitem{tan2020efficientnet}
Mingxing Tan and Quoc~V. Le.
\newblock Efficientnet: Rethinking model scaling for convolutional neural
  networks, 2020.

\bibitem{taskynov2021tensor}
Anuar Taskynov, Vladimir Korviakov, Ivan Mazurenko, and Yepan Xiong.
\newblock Tensor yard: One-shot algorithm of hardware-friendly tensor-train
  decomposition for convolutional neural networks, 2021.

\bibitem{wang2021sampleefficient}
Linnan Wang, Saining Xie, Teng Li, Rodrigo Fonseca, and Yuandong Tian.
\newblock Sample-efficient neural architecture search by learning action space,
  2021.

\bibitem{zhang2017deep}
Ying Zhang, Tao Xiang, Timothy~M. Hospedales, and Huchuan Lu.
\newblock Deep mutual learning, 2017.

\bibitem{zoph2016RL}
Barret Zoph and Quoc~V. Le.
\newblock Neural architecture search with reinforcement learning, 2017.

\bibitem{ZophNasNet}
Barret Zoph, Vijay Vasudevan, Jonathon Shlens, and Quoc~V. Le.
\newblock Learning transferable architectures for scalable image recognition,
  2018.

\end{thebibliography}

\clearpage

\appendix

\section{Transfer learning experimental setup}

For all classification tasks we used the following fine-tuning setup: SGD optimizer with momentum 0.9; learning rate 0.02; 300 epochs with exponential learning rate decay, multiplying it by 0.97 every 2.4 epochs; weight decay 0.0001; 8 NVidia V100 GPUs, total batch size 512; RandAugment augmentation policy; Exponential Moving Average of model weights with coefficient 0.9999.

For the object detection on Pascal VOC with YOLOv3 we used the following setup: image size 608x608, batch size 64 per GPU, 2 NVidia V100 GPU, 52000 iterations of SGD optimizer starting with learning rate 0.0005. We change the learning rate to 0.0001, 0.0002, 0.0005, 0.001, 0.0001 and 0.00001 at steps 400, 700, 900, 1000, 40000 and 45000.

For the object detection on COCO with Faster R-CNN we used the following setup: image size 608x608, batch size 32 per GPU, 8 NVidia V100 GPU, 120000 iterations of SGD optimizer starting with learning rate 0.0001. We change the learning rate to 0.01, 0.001, 0.0001 at steps 5000, 60000 and 80000.

\section{Ablation study for the MEM model}

To study the impact of the latency regression model to $MEM$ value we trained several classic regression models to predict the latency based on the number of matrix operations, vector operations and amount of data transfer for each architecture. The results are presented in the table~\ref{table:mem_ablation}. 

Let $f$ is a regression model and $A$ is an arbitrary architecture with number of matrix, vector and data transfer operations $m(A)$, $v(A)$ and $d(A)$ respectively. Then $$latency(A) = f(m(A), v(A), d(A))$$ 

According to the table~\ref{table:mem_ablation}, all models show similar {\it $R^2$ score} and mean average percentage error ({\it MAPE}), but the {\it linear  regression} model has better or equal quality of the prediction and has the interpretable coefficients which is more important for our study. Thus, we made a decision to use {\it linear  regression} to estimate $MEM$. Note that we didn't consider such models as {\it nearest neighbors regression} and {\it decision trees regression} because these models weren't clearly interpretable in our task. It is important to note that values of $MEM$ score for different architectures are consistent across different regression algorithms.  

\begin{table}[h!]
\centering
\small
\begin{tabular}{|c|c|c|c|c|c|c|}
\hline
\multirow{2}{*}{Model} & \multirow{2}{*}{$R^2$ score} & \multirow{2}{*}{MAPE} & \multicolumn{4}{c|}{MEM Score}  \cr
\cline{4-7}
&&& \rotatebox[origin=c]{90}{~~MobileNetV2~~} & \rotatebox[origin=c]{90}{MNasNet} & \rotatebox[origin=c]{90}{ResNet} & \rotatebox[origin=c]{90}{ISyNet} \cr
\hline
Linear Regression & 0.944 & 12.54\% & 0.167 & 0.17 & 0.294 & 0.378\cr
\hline
Ridge Regression & 0.943 & 12.67\% & 0.144 & 0.148 & 0.266 & 0.349 \cr
\hline
Orthogonal Matching Pursuit~\cite{mallat_omp} & 0.944 & 12.54\% & 0.167 & 0.17 & 0.294 & 0.378\cr
\hline
Bayesian Ridge Regression & 0.944 & 12.54\% & 0.167 & 0.17 & 0.294 & 0.378\cr
\hline
SGD Regressor & 0.942 & 12.74\% & 0.139 & 0.143 & 0.257 & 0.339\cr
\hline
Linear SVR & 0.944 & 12.59\% & 0.155 & 0.158 & 0.279 & 0.363\cr
\hline
\end{tabular}
\caption{Results of different latency prediction models. Linear Regression, Ridge Regression and Bayesian Ridge Regression models were used with default hyperparameters. SGD Regressor and Linear SVR models were used with {\it squared epsilon insensitive} loss, Orthogonal Matching Pursuit was used with {\it number of nonzero coefficients} is equal to 3.}
\label{table:mem_ablation}
\end{table}

\begin{figure}[h!]
\begin{center}
\includegraphics[width=1.0\linewidth]{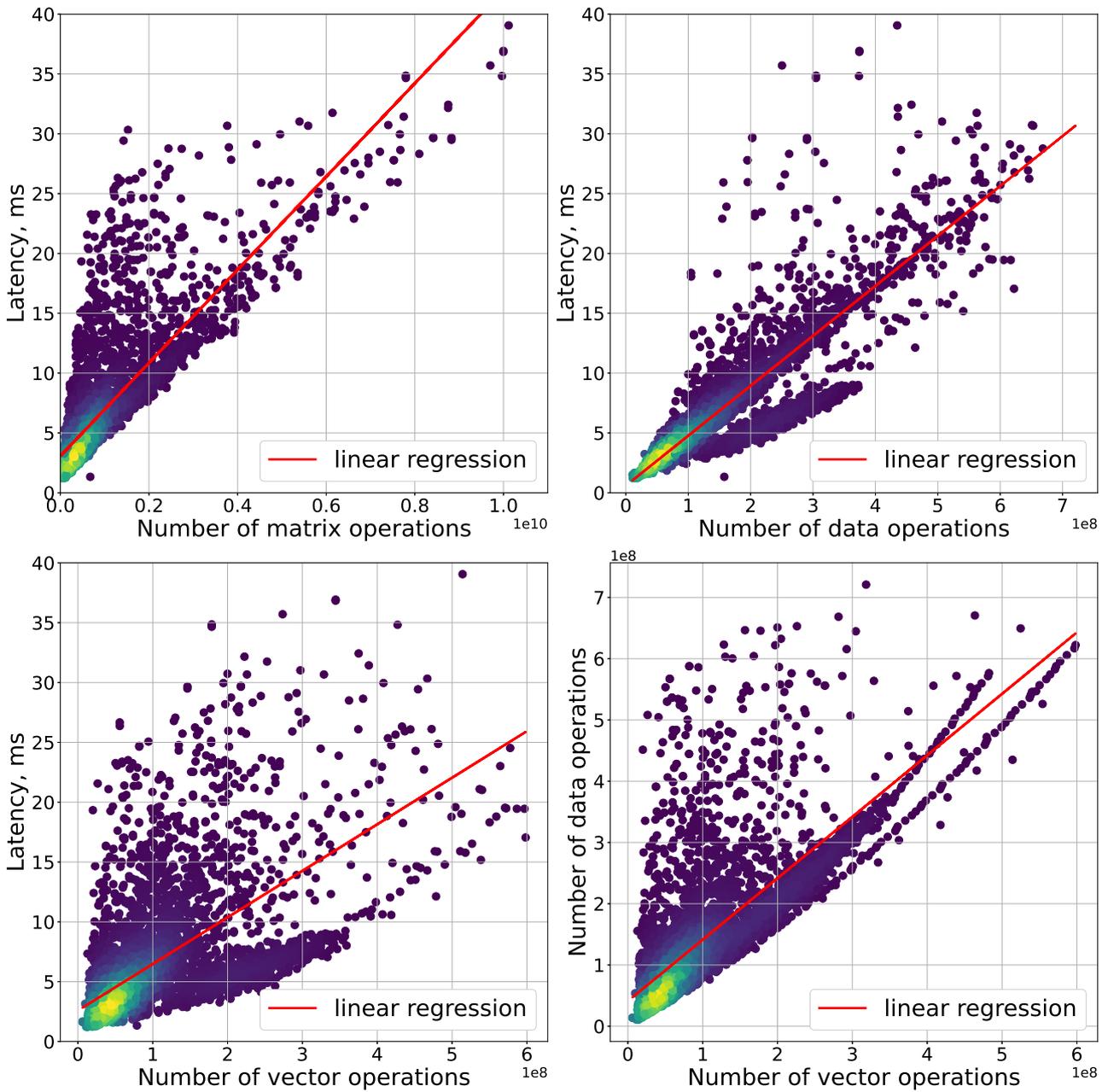}
\end{center}
\caption{Dependency of latency on different types of operations. These plots show how the latency depends on the number of matrix, vector and data transfer operations. Additionally, we show the dependency between number of vector operations and data transfer.}
\label{fig:latency_ops}
\end{figure}

Dependency of latency on different types of operations is shown on the Figure~\ref{fig:latency_ops}.
For some plots (e.g. latency vs. number of memory operations) near-linear dependency exists. For other plots linearity is not so clear, but the linear regression is still correspond to the maximal density of the data points.

\section{Ablation study for the training procedure} \label{section:ablation}

To investigate the impact of each training trick we sequentially disable them and train model ISyNet-N0. Results are shown in the table~\ref{table:tricks_impact}.
We have found that the most beneficial tricks are BatchNorm after classifier's Fully Connected layer;
Exponential Moving Averege~\cite{izmailov2019averaging} on model parameters;
Deep Mutual Learning with stronger peer model~\cite{zhang2017deep};
and training for longer time with smooth decay of learning rate (we multiply it by 0.97 every 2.4 epochs similar to~\cite{tan2019mnasnet}); All the listed tricks improve ISyNet-N3 by 4.69 top-1 ImageNet accuracy.

\begin{table*}[h!]
\centering
\small
\begin{tabular}{|c|c|c|c|c|c|c|c|c|c|c|}
\hline
\multirow{2}{*}{\#} & \multicolumn{6}{c|}{Tricks} & \multirow{2}{*}{Top-1 Acc.} & \multirow{2}{*}{$\Delta$} \cr
\cline{2-7}
& LS & AW & LBN & RA & LT & DML &  & \cr
\hline\hline
1 & - & - & - & - & - & - &  75.92 & \cr 
\hline
2 & - & - & - & - & - & + & 76.77 & +0.85\cr 
\hline
3 & - & - & - & - & + & + & 78.4 & +1.63\cr 
\hline
4 & - & - & - & + & + & + & 79.53 & +1.13\cr 
\hline
5 & - & - & + & + & + & + & 80.08 & +0.55\cr 
\hline
6 & - & + & + & + & + & + & 80.5 & +0.42\cr 
\hline
7 & + & + & + & + & + & + & 80.61 & +0.11\cr 
\hline
\end{tabular}
\caption{The impact of training tricks to the accuracy of ISyNet-N3.
ML denotes mutual learning~\cite{zhang2017deep};
RA denotes RandAugment~\cite{cubuk2019randaugment};
LBN denotes Batch Normalization~\cite{batchnorm} after classifier's fully connected layer;
LT denotes number of training epochs, where '-' is 90 epochs and '+' is 550 epochs;
AW denotes AdamW optimizer~\cite{https://doi.org/10.48550/arxiv.1711.05101};
LS denotes Label Smoothing regularization~\cite{7780677}
}
\label{table:tricks_impact}
\end{table*}

\begin{table*}[h!]
\centering
\small
\begin{tabular}{|c|c|c|c|c|c|c|c|c|c|c|}
\hline
\multirow{2}{*}{\#} & \multicolumn{4}{c|}{Tricks} & \multirow{2}{*}{Top-1 Acc.} & \multirow{2}{*}{$\Delta$} \cr
\cline{2-5}
& RA & DML & LT & LBN &  & \cr
\hline\hline
1  & - & - & - & - &  68.89 & \cr 
\hline
2 & - & - & - & + & 71.59 & +2.7\cr 
\hline
3 & - & - & + & + & 73.11 & +1.52\cr 
\hline
4 & - & + & + & + & 75.11 & +2.0\cr 
\hline
5 & + & + & + & + & 75.32 & +0.21\cr 
\hline
\end{tabular}
\caption{The impact of training tricks to the accuracy of ISyNet-N0.
ML denotes mutual learning~\cite{zhang2017deep};
RA denotes RandAugment~\cite{cubuk2019randaugment};
LBN denotes Batch Normalization~\cite{batchnorm} after classifier's fully connected layer;
LT denotes number of training epochs, where '-' is 90 epochs and '+' is 550 epochs;
}
\label{table:tricks_impact2}
\end{table*}

\end{document}